\documentclass[letterpaper]{article} % DO NOT CHANGE THIS
\usepackage{aaai2026}  % DO NOT CHANGE THIS
\usepackage{times}  % DO NOT CHANGE THIS
\usepackage{helvet}  % DO NOT CHANGE THIS
\usepackage{courier}  % DO NOT CHANGE THIS
\usepackage[hyphens]{url}  % DO NOT CHANGE THIS
\usepackage{graphicx} % DO NOT CHANGE THIS
\usepackage{natbib}  % DO NOT CHANGE THIS AND DO NOT ADD ANY OPTIONS TO IT
\usepackage{caption} % DO NOT CHANGE THIS AND DO NOT ADD ANY OPTIONS TO IT
\usepackage{algorithm}
\usepackage{algorithmic}

\newcounter{procedure}
\newcommand{\PROCEDURE}{\refstepcounter{procedure}\par\vspace{5pt}\noindent\textbf{Procedure \theprocedure}}
\newcommand{\ENDPROCEDURE}{\par\vspace{5pt}}
\usepackage{newfloat}
\usepackage{listings}
\DeclareCaptionStyle{ruled}{labelfont=normalfont,labelsep=colon,strut=off} % DO NOT CHANGE THIS
\floatstyle{ruled}
\newfloat{listing}{tb}{lst}{}
\floatname{listing}{Listing}
\usepackage{booktabs}   % \toprule, \midrule, \cmidrule
\usepackage{multirow}   % only if you later need \multirow
\usepackage{adjustbox}  % convenient, class-safe way to scale wide tables
\usepackage{amsmath}   % for aligned, text, etc.
\usepackage{amssymb}   % optional, but handy
\usepackage[most]{tcolorbox} % loads skins, breakable, listings, etc.
\tcbset{                                       % global defaults
  colback=orange!3!white,
  colframe=black,
  boxrule=0.4pt,
  left=2pt,right=2pt,top=2pt,bottom=2pt,
  enhanced,
  breakable,                                   % page/column breaks
  width=\columnwidth,                          % fit current column
}
\newtcolorbox{promptbox}{}
\usepackage{tabularx}     
\usepackage[table]{xcolor}
\usepackage{amsmath}
\usepackage{enumitem}

\title{PASS: Probabilistic Agentic Supernet Sampling for\\
Interpretable and Adaptive Chest X-Ray Reasoning}

\author{
  Yushi Feng\textsuperscript{\rm 1},
  Junye Du\textsuperscript{\rm 1},
  Yingying Hong\textsuperscript{\rm 1},
  Qifan Wang\textsuperscript{\rm 2},
  Lequan Yu\textsuperscript{\rm 1}\thanks{Corresponding author}
}

\affiliations{
  \textsuperscript{\rm 1} School of Computing and Data Science, The University of Hong Kong, Hong Kong SAR, China\\
  \textsuperscript{\rm 2} Faculty of Engineering, The University of Hong Kong, Hong Kong SAR, China\\
  \{fengys@connect., junyedu@connect., yyhong@, wqf040701@connect., lqyu@\}hku.hk
}

\usepackage{bibentry}
\begin{document}

\maketitle

\begin{abstract}
Existing tool-augmented agentic systems are limited in the real world by (i) black-box reasoning steps that undermine trust of decision-making and pose safety risks, (ii) poor multimodal integration, which is inherently critical for healthcare tasks, and (iii) rigid and computationally inefficient agentic pipelines.
We introduce \textbf{PASS} (\textbf{P}robabilistic \textbf{A}gentic \textbf{S}upernet \textbf{S}ampling), the first multimodal framework to address these challenges in the context of Chest X-Ray (CXR) reasoning. PASS adaptively samples agentic workflows over a multi-tool graph, yielding decision paths annotated with interpretable probabilities.
Given the complex CXR reasoning task with multimodal medical data, PASS leverages its learned task-conditioned distribution over the agentic supernet. Thus, it adaptively selects the most suitable tool at each supernet layer, offering probability-annotated trajectories for post-hoc audits and directly enhancing medical AI safety. PASS also continuously compresses salient findings into an evolving personalized memory, while dynamically deciding whether to deepen its reasoning path or invoke an early exit for efficiency. 
To optimize a Pareto frontier balancing performance and cost, we design a novel three-stage training procedure, including expert knowledge warm-up, contrastive path-ranking, and cost-aware reinforcement learning.
To facilitate rigorous evaluation, we introduce CAB-E, a comprehensive benchmark for multi-step, safety-critical, free-form CXR reasoning.
Experiments across various benchmarks validate that PASS significantly outperforms strong baselines in multiple metrics (e.g., accuracy, LLM-Judge, semantic similarity, etc.) while balancing computational costs, pushing a new paradigm shift towards interpretable, adaptive, and multimodal medical agentic systems.
\end{abstract}

%----------------------
\section{Introduction}
%----------------------

\begin{figure*}[t]
    \centering
    \includegraphics[width=1\textwidth]{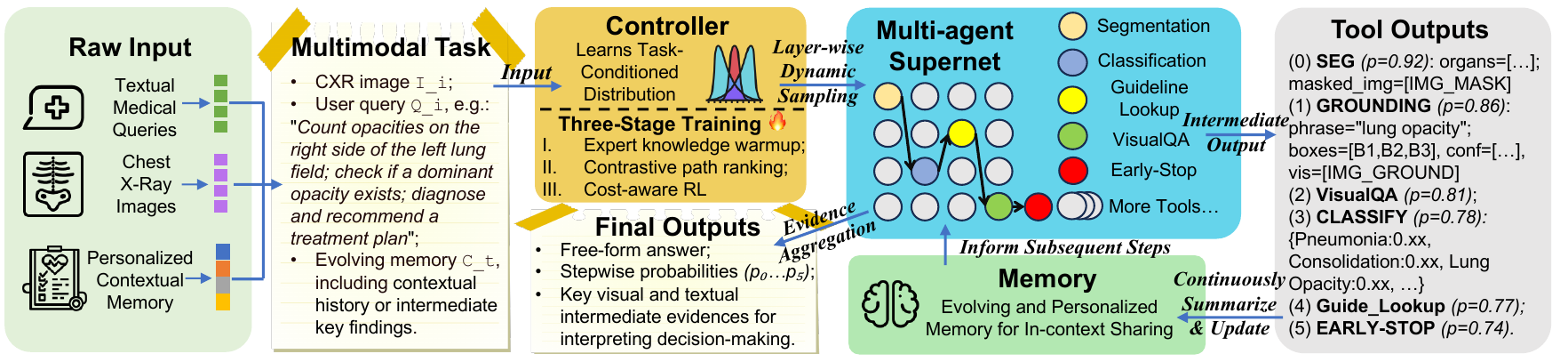}
    \caption{An overview of PASS. Given a multimodal complex reasoning task (CXR image, textual comprehensive query, multimodal personalized context), our probabilistic controller learns a continuous task-conditioned distribution over the agentic supernet (i.e. a directed acyclic graph of medical agent containers). At each step, it samples an action, yielding a workflow annotated with interpretable probabilities for post-audits and directly enhances clinical AI safety. Tool outputs, which can be both text and images, are summarized and fed into an evolving personalized memory and shared in-context to inform subsequent steps. The controller is trained via a principled three-stage strategy (expert knowledge warm-up, contrastive path ranking, cost-aware reinforcement learning) to optimize the accuracy-cost trade-off. Eventually, PASS is enabled to answer multimodal medical questions in free-form text via an interpretable, adaptive, and efficient agentic reasoning process.}
    \label{fig:framework}
\end{figure*}

Chest X-Ray is the most commonly performed diagnostic imaging procedure worldwide, widely regarded as a cornerstone of modern radiology
~\cite{johnson2019mimic}. 
However, interpreting CXRs demands careful multi-structure assessment that is time-consuming and expertise-intensive~\cite{bahl2020}. 
While specialized AI tools for tasks like classification~\cite{chexnet2017}, segmentation~\cite{medsam2024} or report generation~\cite{tanno2024,chexpertplus2024} etc. have shown promise in improving turnaround time and diagnostic consistency~\cite{baltruschat2021,ahn2022association,pham2022,shin2023}, their narrow specialization hinder their use in complex clinical reasoning scenarios~\cite{erdal2023,fallahpour2024}.

Large-scale foundation models (FMs) in recent years like GPT-4o~\cite{openai2024gpt4ocard}, LLaVA-Med~\cite{li2023llava}, and CheXagent~\cite{cxragent2024} offer a more unified approach by integrating visual and textual reasoning. However, these monolithic systems often hallucinate~\cite{eriksen2024use}, lack domain-specific robustness~\cite{cxragent2024}, and operate as uninterpretable ``black boxes", making them unsuitable for high-stakes medical deployment.

Motivated by the need for more reliable, generalized, and autonomous solutions, recent efforts have explored \emph{multi-agent medical AI systems} that coordinate domain-specific tools utilizing the capability of large language models (LLMs) and vision language models (VLMs). 
Recent progress in general-purpose agent systems~\cite{camel2023,autogen2023,zhuge2024gptswarm} demonstrate the potential of collaborative LLM agents to outperform single-agent baselines through structured communication and role specialization~\cite{du2023,liang2024}. Despite these advances, most systems rely on manually-defined and rigid workflows~\cite{qian2024,zhang2025aflow}, which cannot adapt to the varying complexity of clinical queries and are computationally inefficient.

To address these challenges, recent methods have aimed to automate the design of multi-agent workflows. 
Works such as DsPy~\cite{khattab2024} and EvoPrompt~\cite{guo2024} optimize prompts, while G-Designer~\cite{gdesigner2024} and AutoAgents~\cite{chen2024autoagents} refine inter-agent communication and profiling strategies. 
In the medical domain, MedRAX~\cite{fallahpour2025medrax} exemplifies this direction by orchestrating multiple CXR tools via ReAct-style prompting \cite{yao2023react}, achieving improved accuracy over end-to-end models. However, these methods largely rely on black-box LLMs for the decision-making of invoking agents, leaving the concerns regarding trustworthiness and safety risks as open questions.

The most recent advance, agentic supernets like MaAS~\cite{maas2025}, introduced a paradigm shift by learning a distribution over possible workflows, enabling adaptive, cost-aware reasoning.
However, this approach has two fundamental flaws for medical applications. First, it is designed for text-only reasoning and lacks multimodal integration, which is inherently a core requirement in clinical reasoning. Second, while its textual gradient mechanism enables workflow optimization, it operates implicitly within the LLM’s internal prompt space during multi-turn conversations, providing limited interpretability and traceability in high-stakes use.

These challenges highlight a critical need for a medical agentic system that is not only multimodal and truly interpretable, but also adaptive and efficient.
To this end, we propose \textbf{PASS} (\textbf{P}robabilistic \textbf{A}gentic \textbf{S}upernet \textbf{S}ampling). To the best of our knowledge, PASS is the \textit{first} framework for interpretable and adaptive CXR reasoning via multimodal agentic workflow sampling. 
Given a CXR image and a complex free-form clinical reasoning task, PASS manages an evolving contextual memory, operates over a directed acyclic graph consisting of multiple specialized medical agent containers (i.e., agentic supernet), and adaptively samples layer-wise tool sequences from the graph. Crucially, we design a Controller module to learn the task-conditioned continuous distribution over the supernet, yielding decision paths annotated with interpretable probabilities. This provides transparent trajectories for post-hoc audits, directly enhancing medical AI safety. 
We design a principled three-stage regimen for the training of PASS: (1) expert knowledge-guided warm-up aligns tool usage with clinical best practices; (2) contrastive path-ranking sharpens ordering preferences among tool sequences; and (3) cost-aware reinforcement learning trains the controller to learn the optimized accuracy-cost Pareto frontier with an early-exit mechanism.

To systematically evaluate such agentic systems, where existing CXR benchmarks largely focus on simplified classification or short-form QA and are thus poorly aligned with this paradigm, we introduce \textsc{ChestAgentBench-E} (CAB-E), a new challenging new benchmark comprising 2,550 comprehensive and safety-critical CXR reasoning cases annotated with free-form QA pairs, image inputs, and queries that demand highly complex rationales\footnote{Code, data, and benchmarks are available for research purposes at \url{https://github.com/ys-feng/PASS}.}. 
CAB-E expands the scope of prior evaluations~\cite{fallahpour2025medrax,shao2021slake}, emphasizing multi-step and clinically grounded queries that require adaptive tool orchestration. It also evaluates free-form answering and safety-critical cases.

Our key contributions can be summarized as follows:
\begin{itemize}
  \item We propose PASS, the first framework to our knowledge to instantiate a probabilistic agentic supernet for multimodal medical reasoning, representing a paradigm shift towards building trustworthy, adaptive, transparent, and cost-aware agentic systems.
  \item We design a principled three-stage training strategy including expert knowledge guided warm-up, contrastive path ranking, and cost-aware reinforcement learning.
  \item We introduce \textsc{CAB-E}, a comprehensive public benchmark to evaluate multi-hop and safety-critical agentic reasoning for CXR with free-form answers.
  \item Extensive experiments validate that PASS outperforms strong baselines among various benchmarks, while maintaining the balanced computational cost and providing interpretable agentic workflows.
\end{itemize}

%===========================================================
\section{Methodology}
\label{sec:method}
%===========================================================

In this paper, we propose a probabilistic framework for \textbf{PASS} that interprets workflow construction as a latent decision-making process governed by a multimodal generative policy. In this section, we first formulate a probabilistic controller over tool trajectories and answers, and derive a cost-aware objective grounded in expected utility maximization. We then introduce the architecture and parameterization of the controller $\pi_\theta$, followed by a theoretically motivated multi-phase training algorithm that combines expert knowledge warm-up, contrastive path ranking and cost-aware reinforcement learning.

\subsection{Preliminary}
\paragraph{Problem formulation and notations.}
Let $\mathcal{Q}=\{(q_i,I_i,C_i)\}_{i=1}^{N}$ be a collection of {\it multimodal diagnostic queries}, where $q_i\!\in\!\mathcal{T}$ is a free-form text question, $I_i\!\in\!\mathbb{R}^{H\times W\times3}$ is a chest X-ray image and $C_i\!\in\!\mathcal{C}$ denotes personalized contextual memory, including summarized information like structured demographic factors, clinical results, and previous analysis outputs.
PASS answers $q_i$ by sampling a {\it workflow} $\tau$ over a directed acyclic \emph{multi-container graph} and executing the tools in corresponding containers in the selected sequence. 
We frame workflow generation as sampling from a probability distribution $\pi_\theta$ based on multimodal evidence:
\begin{equation}
    \tau \sim \pi_\theta(\,\cdot \mid q,I,C), \quad
    \hat a=\textsc{Execute}(\tau)
\end{equation}
where the {\it workflow} $\tau = (a_1,a_2,\ldots,a_T)$ is the trajectory of the actions and $\hat a$ is a free-form answer (e.g., finding, measurement, report section, etc.) returned to the clinician.  PASS must simultaneously maximize diagnostic utility $\mathcal{U}$ and minimize a composite cost $\mathcal{L}$ capturing latency, token usage and privacy risk. The hyperparameter $\lambda$, configured by the user or deploying institution, controls the trade-off between performance and operational constraints. Under the above settings, the goal of our model could be stated as:

\begin{equation}
\max_{\theta}\;
\mathbb{E}_{\substack{(q,I,C)\sim\mathcal{Q} \\ \tau\sim\pi_\theta}}
\Bigl[
\mathcal{U}(\hat{a}, a^\star) - \lambda \cdot \mathcal{L}(\tau)
\Bigr]
\label{eq:overall_obj}
\end{equation}

\paragraph{Agentic supernet.}
Supernet $\mathcal{G}=(\mathcal{V},\mathcal{E})$ contains agent containers as nodes and legal tool invocations as edges. Each container $v\!\in\!\mathcal{V}$ is typed by one of  
\textsc{Segmentation},\;\textsc{Classify},\;\textsc{Grounding},\;\textsc{Report},\;
\textsc{VQAnalyze},\;\textsc{GuidelineLookup} and \textsc{MKG}.
The container $v$ also stores a mutable set of tool models $T_v=\{t_{v,1},\dots\}$ that share identical I/O signature but may differ in backbone architecture, patch size or training epoch. The detailed tool model descriptions are in the Appendix\footnote{Appendices are available at \url{https://arxiv.org/abs/2508.10501}}.
Edges $e=(v\!\rightarrow\!v')\!\in\!\mathcal{E}$ are labeled with a routing policy $\rho_{e}$ specifying which fields of the current memory are forwarded to the next container.

\begin{algorithm}[t]
\caption{\textsc{PASS}: Training Procedure}
\label{alg:passpp-train}
\small
\begin{algorithmic}[1]
\REQUIRE Expert demonstrations $\mathcal{D}_{\mathrm{exp}}$, unlabeled data $\mathcal{D}_{\mathrm{ul}}$, supernet $\mathcal{G}$, state encoder $\psi$, policy $\pi_\theta$, answer generator $p_\phi$, heuristic reward $R_h$, cost weights $\lambda$, entropy weight $\gamma$.

\PROCEDURE
  \STATE {\textit{\# Phase I: Expert Knowledge Warm-up}}
  \FOR{$(s, a^\star) \in \mathcal{D}_{\mathrm{exp}}$}
    \STATE $\theta \leftarrow \theta - \eta_1 \nabla_\theta \bigl( -\log \pi_\theta(a^\star \mid s) \bigr)$
  \ENDFOR
  \STATE {\textit{\# Phase II: Heuristic-Guided Path Ranking}}
  \FOR{$(q, I, C) \in \mathcal{D}_{\mathrm{ul}}$}
    \STATE $\{\tau_k\}_{k=1}^{K} \sim \pi_\theta(\cdot \mid q, I, C)$
    \STATE $p(\tau_k) \leftarrow \frac{\exp(R_h(\tau_k)/\alpha_{\mathrm{cpr}})}{\sum_{j=1}^{K} \exp(R_h(\tau_j)/\alpha_{\mathrm{cpr}})}$
    \STATE $\mathcal{L}_{\textsc{CPR}} \leftarrow - \sum_{k=1}^{K} p(\tau_k) \log \pi_\theta(\tau_k)$
    \STATE \textit{Update $\theta$ using} $\nabla_\theta \mathcal{L}_{\textsc{CPR}}$
  \ENDFOR
  \STATE {\textit{\# Phase III: Cost-aware Reinforcement Learning}}
  \FOR{$n = 1$ \TO $N_{\mathrm{RL}}$}
    \STATE $\tau \sim \pi_\theta(\cdot \mid q, I, C) \in \mathcal{D}_{\mathrm{ul}}$
    \STATE $\hat{a} \sim p_\phi(\cdot \mid \tau, q, I, C)$
    \STATE $R(\tau) \leftarrow \mathcal{U}(\hat{a}, a^\star) - \lambda \mathcal{L}(\tau) - \gamma H(\hat{a})$
    \STATE $\theta \leftarrow \theta + \eta_3 R(\tau) \nabla_\theta \log \pi_\theta(\tau)$
  \ENDFOR
\ENDPROCEDURE
\end{algorithmic}
\end{algorithm}

\begin{algorithm}[t]
\caption{\textsc{PASS}: Inference}
\label{alg:passpp-infer}
\small
\begin{algorithmic}[1]
\REQUIRE Policy $\pi_\theta$, generator $p_\phi$, summarizer $\mathcal{S}$, state encoder $\psi$, supernet $\mathcal{G}$, max steps $T_{\max}$.
\PROCEDURE
    \STATE $M, \tau \leftarrow \emptyset, []$  \quad \textit{// Initialize memory and trajectory}
    \FOR{$t = 1$ \TO $T_{\max}$}
      \STATE $a_t \sim \pi_\theta(\cdot \mid \psi(q, I, C, M))$
      \IF{$a_t = \textsc{EarlyExit}$}
        \STATE \textbf{break}
      \ENDIF
      \STATE $\rho_t \leftarrow \textsc{ExecuteTool}(a_t)$
      \STATE $M \leftarrow M \cup \mathcal{S}(\rho_t)$
      \STATE $\tau \leftarrow \tau \cdot a_t$
    \ENDFOR
    \STATE $\hat{a} \sim p_\phi(\cdot \mid q, I, C, \tau)$
    \STATE \textsc{Return} $(\hat{a}, \tau)$ \quad \textit{// Return final answer and full workflow}
\ENDPROCEDURE
\end{algorithmic}
\end{algorithm}

\paragraph{Formal Interface.}
Every container $v$ adheres to a unified formal interface, defining its input $x_v$ and output $y_v$ as:
\begin{equation}
\left\{
\begin{aligned}
    x_v &= \bigl(q^{(\mathrm{sub})},\, I^{(\mathrm{roi})},\, C^{(\mathrm{sub})},\, \eta\bigr) \\
    y_v &= \bigl(\rho_v,\, \ell_v,\, \kappa_v\bigr)
\end{aligned}
\right.
\end{equation}
\noindent The input consists of a textual sub-query $q^{(\mathrm{sub})}$, an optional region-of-interest image tensor $I^{(\mathrm{roi})}$, a relevant slice of personalized contextual memory $C^{(\mathrm{sub})}$, and tool-specific hyperparameters $\eta$. The output comprises the \emph{primary multimodal payload} $\rho_v$—which may be a JSON object for structured data (e.g., \textsc{{text, bbox, prob}}) or an image tensor for visual data (e.g., a segmentation mask)—along with the measured \emph{latency} $\ell_v$ and \emph{token cost} $\kappa_v$, both utilized in the overall objective (Eq.~\eqref{eq:overall_obj}). By strictly enforcing this interface across all containers, our design ensures seamless plug-and-play integration and maintenance.
\paragraph{Action space.}
The space spanned by legal actions at state $s_t$ is defined as:
\[
    \mathcal{A}(s_t)=
    \bigl\{(v,T_{v,k})\mid (v_t\!\rightarrow\!v)\in\mathcal{E}\bigr\}
    \cup\{\textsc{EarlyExit}\}
\]
\noindent where $(v,T_{v,k})$ denotes executing tool $T_{v,k}$ inside container $v$.  Sampling the \textsc{EarlyExit} action, a special action that halts the execution trajectory early to conserve resources, thus initiating answer synthesis in advance.

\subsection{Multi-agent Workflows}
\label{subsec:gen-model}
PASS models diagnostic reasoning as structured decision-making in a latent space of tool-based workflows. Given an input triplet $(q,I,C)$, the agent sequentially builds a trajectory $\tau = (a_1,a_2,\ldots,a_T)$ by sampling actions $a_t \in \mathcal{A}(s_t)$, where $s_t$ is the multimodal reasoning state at step $t$. The agent's final output is a multimodal package, consisting of a final textual answer $\hat{a} \in \mathcal{T}$ and any visual artifacts (e.g., annotated images) produced during the workflow $\tau$.

The core of PASS is the workflow policy $\pi_\theta(a_t \mid s_t)$, which we aim to learn. This policy, combined with a fixed answer synthesis module $p_\phi$, defines the full generative process for the textual answer $\hat{a}$:
\begin{equation}
\begin{aligned}
    &p_\theta(\tau, \hat{a} \mid q, I, C)  \\
    =&  \underbrace{p_\phi(\hat{a} \mid \tau, q, I, C)}_{\text{Answer generator}} \cdot \underbrace{\prod_{t=1}^{T} \pi_\theta(a_t \mid s_t)}_{\text{Workflow policy}} 
    \label{eq:joint}
\end{aligned}
\end{equation}
\noindent where $p_\phi$ is a frozen synthesis module (e.g., a large language model) responsible for generating the final text answer $\hat{a}$ based on the evidence gathered in $\tau$. All learning is concentrated in the policy parameters $\theta$, ensuring improvement stems from discovering better workflow decisions, not from fine-tuning the generator. This decomposition makes two key assumptions: (i) the tool sampling is a Markov process over the state space $\mathcal{S}$, and (ii) the final textual answer is conditionally independent of the internal policy decisions, given the full trajectory $\tau$.

\paragraph{Policy-induced answer distribution.}
By virtue of marginalizing out the latent tool trajectory $\tau$, we obtain the model-implied distribution over answers:
\begin{equation}
\begin{aligned}
    &p_\theta(\hat{a} \mid q, I, C)  \\
    =& \sum_{\tau \in \mathcal{T}(q,I,C)} 
      \pi_\theta(\tau \mid q,I,C) \cdot p_\phi(\hat{a} \mid \tau, q, I, C)
    \label{eq:marginal}
\end{aligned}
\end{equation}
in which $\pi_\theta(\tau \mid q,I,C) = \prod_{t=1}^{|\tau|} \pi_\theta(a_t \mid s_t)$ and $\mathcal{T}(q,I,C)$ is the set of legal trajectories under $\mathcal{G}$ from initial state $s_0$. Although this marginal distribution is intractable to compute exactly due to the combinatorial size of $\mathcal{T}$, it can be approximated with Monte Carlo sampling, which we exploit both for training and for uncertainty estimation.

\paragraph{Expected utility and cost regularization.}
Given a ground-truth answer $a^\star$ and a reward function $\mathcal{U}(\hat{a}, a^\star)$ measuring the clinical utility of the predicted answer, our goal is to maximize the expected utility of our policy. This objective must also be balanced against the cost of the workflows it generates. Formally, the goal is to find optimal parameters $\theta$ for the policy $\pi_\theta$:
\begin{align}
\max_\theta \;\; \mathbb{E}_{(q,I,C)\sim\mathcal{Q}} \Bigl[ \mathbb{E}_{\hat{a} \sim p_\theta(\cdot \mid q,I,C)} \mathcal{U}(\hat{a}, a^\star) \notag \\
- \lambda \cdot \mathbb{E}_{\tau \sim \pi_\theta(\cdot \mid q,I,C)} \mathcal{L}(\tau) \Bigr]
\label{eq:expected-obj}
\end{align}
\noindent This formulation can be viewed as a constrained variational inference problem over the latent workflow $\tau$ with an amortized inference network $\pi_\theta$.

\paragraph{Uncertainty-aware generation.}
The posterior entropy of the answer distribution, $H_\theta(\hat{a} \mid q,I,C) = -\mathbb{E}_{\hat{a}} \log p_\theta(\hat{a} \mid q,I,C)$, can be utilized to quantify the epistemic uncertainty of the model. Since the answer generator $p_\phi$ is frozen, this entropy is solely induced by the sampling variability in the workflow trajectory $\tau \sim \pi_\theta$. In practice, we estimate $H_\theta$ via Monte Carlo rollouts of the policy and use it both as a proxy for answer confidence and as a regulariser during policy learning (Sec.~\ref{sec:training}) to discourage high-entropy outputs in high-risk settings.

%-----------------------------------------------------------
\subsection{Controller Architecture}
\label{subsec:controller}
%-----------------------------------------------------------

The controller \(\pi_\theta(a_t \mid s_t)\) is designed as a masked categorical distribution over permissible actions, with its parameters determined by a state encoder \(\psi\).
%and a multilayer perceptron head. 
Its logits are produced by a policy network head that processes the state representation \(h_t\).
Let $s_t=(q,I,C,M_t)$ denote the current multimodal state. The state encoder maps this input into a shared representation $h_t \in \mathbb{R}^{d}$:
\begin{equation}
\begin{gathered}
h_t = \psi(s_t) = \mathrm{LN}(z_t) \quad s.t.\\
z_t = W_I \cdot \xi(I) \;\Vert\; W_Q \cdot \zeta(q,C) \;\Vert\; W_M \cdot \mu(M_t)
\end{gathered}
\label{eq:encoder}
\end{equation}

\noindent where $\xi(I)$ is a frozen ViT-B/16 image encoder with final-layer CLS token projected to $\mathbb{R}^{256}$,
$\zeta(q,C)$ is a Sentence-BERT-style text encoder for $(q,C)$, projected to $\mathbb{R}^{128}$,
$\mu(M_t)$ encodes dynamically updating memory over its summaries, with pooled final hidden state $\in \mathbb{R}^{128}$,
$\mathrm{LN}(\cdot)$ denotes layer normalization, and $\Vert$ denotes concatenation. 
The policy head is a feed-forward network with a single hidden layer and ReLU activation:
\begin{equation}
\begin{aligned}
    &\pi_\theta(a_t \mid s_t)  \\
    =& \mathrm{Softmax}\left( \mathrm{mask}_{\mathcal{A}(s_t)} \left[
    W_2 \cdot \sigma(W_1 h_t) \right] \,/\, \alpha \right)
    \label{eq:controller-head}
\end{aligned}
\end{equation}

\noindent where the legal-action mask $\mathrm{mask}_{\mathcal{A}(s_t)}$ zeroes out infeasible transitions in the supernet $\mathcal{G}$ and $\alpha$ is a temperature parameter annealed during training from 2.0 to 0.8.

\paragraph{Personalized contextual memory.}
At step $t$, what the controller observes are stated as:
\[
    s_t=\Bigl(q,I,C,M_t\Bigr) , \quad
    M_t=\bigl\{(v_j,\tilde y_{v_j})\bigr\}_{j=1}^{t-1}
\]

\noindent where the memory $M_t$ is a bounded-size first-in-first-summarized (FIFS) buffer. After each tool call, in order to save the computational cost, the JSON response $y_v$ is summarized to a compressed vector $\tilde y_v$ using a frozen language model prompted to function only as paraphrasing. These textual summaries are appended to a FIFO memory $M_t$ along with image outputs (if any). This personalized and evolving memory mechanism enables precise, in-context diagnosis in the wild.

\begin{table*}[t]
\centering
\small
\setlength{\tabcolsep}{4pt}
\renewcommand{\arraystretch}{1.1}
\begin{adjustbox}{width=\linewidth}
\begin{tabular}{l|ccccccc|cc|cc}
\toprule
\multirow{2}{*}{\textbf{Model}} & \multicolumn{7}{c|}{\textbf{CAB-E}} & \multicolumn{2}{c|}{\textbf{CAB-Standard}} & \multicolumn{2}{c}{\textbf{SLAKE}} \\
\cmidrule(lr){2-8}\cmidrule(lr){9-10}\cmidrule(lr){11-12}
& Acc.$\uparrow$ & LLM-J.$\uparrow$ & BLEU$\uparrow$ & METEOR$\uparrow$ & ROUGE-L$\uparrow$ & Sim.$\uparrow$ & Lat.$\downarrow$ & Acc.$\uparrow$ & Lat.$\downarrow$ & Sim.$\uparrow$ & Lat.$\downarrow$ \\
\midrule
GPT-4o (zero-shot)      &  60.06 $\pm$ 0.01& 45.29 $\pm$ 0.07& 4.09 $\pm$ 0.03& 25.63 $\pm$ 0.02& 25.84 $\pm$ 0.01& 79.03  $\pm$ 0.01& 18,37 & 45.45 $\pm$ 0.02& \underline{3.10} & 37.25 $\pm$ 0.03&  \underline{2.25}\\
\quad CoT              &  59.18       $\pm$ 0.01& 39.43 $\pm$ 0.06& 3.83 $\pm$ 0.03& 23.93 $\pm$ 0.02& 25.25 $\pm$ 0.01& 77.62 $\pm$ 0.01& 20.30 & 50.51 $\pm$ 0.02& 3.34 & 38.78 $\pm$ 0.02& 2.43 \\
\quad ComplexCoT       &  63.26       $\pm$ 0.01& 41.06       $\pm$ 0.06& 4.22       $\pm$ 0.04& 25.14       $\pm$ 0.02& 25.12       $\pm$ 0.02& 78.03       $\pm$ 0.01& 22.17 & 44.44 $\pm$ 0.01& 3.41 & 42.86 $\pm$ 0.03& 2.57 \\
\quad SC (CoT$\times$5) &  79.59     $\pm$ 0.08& 54.13     $\pm$ 0.07& 5.34     $\pm$ 0.01& 31.22     $\pm$ 0.02& 25.83     $\pm$ 0.01& 76.14     $\pm$ 0.03& \underline{14.55} & 43.43 $\pm$ 0.02 & 10.35 & 44.88 $\pm$ 0.02& 7.83 \\
GPT\mbox{-}4o (finetuned) & 81.82 $\pm$  0.06& 75.76 $\pm$ 0.02& \textbf{18.20} $\pm$ 0.01& \underline{32.92} $\pm$ 0.01& 44.49 $\pm$ 0.02& 88.19 $\pm$ 0.01& 14.99 & 62.83 $\pm$ 0.01& 3.79 & 81.82 $\pm$ 0.01& 3.36 \\
o3-mini (+visual tool) & 73.73 $\pm$  0.01& 68.08 $\pm$  0.04&  4.43 $\pm$  0.01& 33.09 $\pm$  0.01& 24.52 $\pm$  0.01& 80.21 $\pm$  0.02& 41.91 & 50.51 $\pm$  0.01& 26.18 & 54.55 $\pm$  0.01& 11.63 \\
CheXagent              &  83.67      $\pm$ 0.01& 69.47      $\pm$ 0.01& 2.71      $\pm$ 0.01& 14.68      $\pm$ 0.01& 20.78      $\pm$ 0.01& 82.52      $\pm$ 0.01& \textbf{2.20}      & 62.63 $\pm$ 0.03 & \textbf{0.40} & \underline{78.80} $\pm$ 0.01 & \textbf{0.65}\\
LLaVA-Med              &  86.96       $\pm$ 0.05& \underline{82.65}       $\pm$ 0.04& 8.28      $\pm$ 0.01& 29.96       $\pm$ 0.01& \underline{31.26}       $\pm$ 0.01& \textbf{91.00}       $\pm$ 0.01& 21.43       & 53.23 $\pm$ 0.01       & 7.79 & 60.60 $\pm$ 0.01& 10.14\\
MedRAX                 &  \underline{89.54}      $\pm$ 0.02& 76.94      $\pm$ 0.01& 5.56      $\pm$ 0.02& 32.84      $\pm$ 0.05& 27.11      $\pm$ 0.02& 88.69      $\pm$ 0.02& 17.44 & \underline{63.49} $\pm$ 0.02 & 7.39 & 74.90 $\pm$ 0.02& 10.47\\
\textbf{PASS (Ours)}   & \textbf{91.22} $\pm$ 0.12& \textbf{84.28} $\pm$ 0.10& \underline{8.51} $\pm$ 0.05& \textbf{33.21} $\pm$ 0.05& \textbf{31.49} $\pm$ 0.09& \underline{90.16} $\pm$ 0.04& 22.06 & \textbf{66.10} $\pm$ 0.03& 8.05 & \textbf{79.55} $\pm$ 0.04& 11.68 \\
\bottomrule
\end{tabular}
\end{adjustbox}
\caption{Performance across three radiology VQA benchmarks (mean $\pm$ standard deviation).   Best and runner-up numbers are bold and underlined. }
\label{tab:main_results}
\end{table*}

%-----------------------------------------------------------
\subsection{Three-Stage Training Procedure}
\label{sec:training}
%-----------------------------------------------------------

We train the workflow policy $\pi_\theta$ to optimize the objective in Eq.~\eqref{eq:expected-obj} via a principled three-stage procedure. This curriculum-based approach progressively refines the policy, starting with strong expert supervision before moving to weaker preference signals and finally to direct reinforcement learning on the end-task reward. The three stages are detailed as follows. Each stage is grounded in a formal objective, allowing for stable and efficient training of $\pi_\theta$.

\paragraph{Phase I: Expert knowledge guided warm-up.}
This initial phase uses imitation learning to bootstrap the policy. We construct a dataset of expert demonstrations, $\mathcal{D}_{\mathrm{exp}}$, not from scratch, but by using a more scalable, two-step process. First, we use a powerful foundation model (GPT-4o) to generate initial workflow sketches for a set of problems. Second, these sketches are then reviewed, corrected, and validated in a human-in-the-loop process by licensed radiologists. This ``distill-and-refine'' strategy yields a high-quality dataset of one-step decisions $\mathcal{D}_{\mathrm{exp}} = \{(s, a^\star)\}$, where $a^\star$ is the expert-verified action for state $s$. We warm-start the policy by minimizing the KL divergence from the expert policy (i.e., behavior cloning):
\begin{equation}
    \mathcal{L}_{\textsc{BC}} = \mathbb{E}_{(s,a^\star) \sim \mathcal{D}_{\mathrm{exp}}} 
    \left[ -\log \pi_\theta(a^\star \mid s) \right]
\end{equation}
This phase instills a strong prior in the policy, anchoring it in clinically valid reasoning patterns.

\paragraph{Phase II: Heuristic-guided contrastive path ranking.}
Expert demonstrations are costly to acquire and cannot cover all scenarios. To generalize beyond $\mathcal{D}_{\mathrm{exp}}$, we introduce a weaker supervisory signal based on heuristic preferences for unlabeled data. For a given query, we sample $K$ candidate workflows $\{\tau_k\}_{k=1}^K$ from the current policy $\pi_\theta$. We then score each path using a heuristic reward function, $R_h(\tau_k)$, which combines domain-specific priors such as clinical guideline compliance, anatomical coherence, and brevity. The policy is then updated using a contrastive loss (InfoNCE) that encourages it to assign higher probability to higher-scoring paths:
\begin{equation}
\begin{split}
\mathcal{L}_{\textsc{CPR}} = &
\mathbb{E}_{\{\tau_k\} \sim \pi_\theta} \left[ - \sum_{k=1}^{K} p(\tau_k) \log \pi_\theta(\tau_k) \right], \\
& \text{where } p(\tau_k) = \frac{\exp(R_h(\tau_k)/\alpha_\mathrm{cpr})}{\sum_{j=1}^{K} \exp(R_h(\tau_j)/\alpha_\mathrm{cpr})}
\end{split}
\end{equation}
where $\alpha_\mathrm{cpr}$ is a temperature hyperparameter. This phase teaches the policy to distinguish between good and bad reasoning structures, even without a ground-truth workflow.

\paragraph{Phase III: Cost-aware reinforcement learning.}
In the final phase, we directly fine-tune the policy $\pi_\theta$ using reinforcement learning to maximize the expected end-task utility. To compute the reward for a generated workflow $\tau$, we first use the fixed answer generator $p_\phi$ to synthesize a textual answer, $\hat{a} \sim p_\phi(\cdot \mid \tau, q, I, C)$. We then define the reward for the trajectory as:
\begin{equation}
    R(\tau) = \mathcal{U}(\hat{a}, a^\star) 
            - \lambda \cdot \mathcal{L}(\tau)
            - \gamma \cdot H(\hat{a})
    \label{eq:pg-reward}
\end{equation}
where $H(\hat{a})$ is the entropy of generated answers, penalizing uncertainty. We then update the policy parameters $\theta$ using a reinforcement learning approach. The objective is to maximize the expected reward over all trajectories sampled from the policy:
\begin{equation}
    J(\theta) = \mathbb{E}_{\tau \sim \pi_\theta} [R(\tau)]
\end{equation}
The gradient of this objective, $\nabla_\theta J(\theta)$, can be estimated using sampling via the reinforcement algorithm, with a baseline to reduce variance. This final tuning step aligns the workflow generation directly with the ultimate goals of diagnostic accuracy and computational efficiency.

%===========================================================
\section{Experiments}
\label{sec:exp}
%===========================================================

We evaluate PASS across three radiology benchmarks of increasing complexity to assess four critical aspects of real-world deployment: clinical accuracy, language fidelity, computational efficiency, and safety. All experiments are conducted on a single NVIDIA H800 (80GB) GPU with access to OpenAI's GPT API for relevant baselines.

\subsection{Experiment Setup}

\noindent\textbf{Benchmarks.}
We use the following evaluation suites, with more details described in the Appendices:
\begin{itemize}
    \item \textbf{SLAKE}~\cite{shao2021slake}: A native free-form medical VQA benchmark with 6,437 image–question pairs, used to assess zero-shot generalization.
    \item \textbf{CAB-Standard}~\cite{fallahpour2025medrax}: A multiple-choice Chest Agent Benchmark (CAB) containing 2,500 diagnostic queries. CAB-Standard is constructed using the generation method proposed by Fallahpour et al.
    \item \textbf{CAB-E:} Our proposed benchmark with 2,550 multi-step CXR reasoning cases, including 500 safety-critical instances. Construction details and summary statistics are provided in Appendices A and F. This benchmark is designed to evaluate free-form, multi-hop reasoning grounded in both imaging data and patient context. The safety-critical subset focuses on complex, high-stakes scenarios that demand careful and transparent decision-making, such as life-threatening anatomical abnormalities and urgent systemic conditions. CAB-E is publicly available at the aforementioned URL.

\end{itemize}

\noindent\textbf{Metrics.}
On CAB-E, we report: Accuracy, LLM-as-a-Judge score (LLM-J.) based on human expert-guided rubrics, BLEU, METEOR, ROUGE-L, embedding similarity, and end-to-end latency. CAB-Standard is evaluated by accuracy and latency. SLAKE is evaluated by embedding similarity and latency. We evaluate the hallucination rate on the safety-critical split of CAB-E, report blind human radiologist evaluation, and compare the inference cost against LLM-J. to assess the models' efficiency. We present detailed descriptions of the metrics in Appendix B.

\begin{table}[t]
\centering\small
\setlength{\tabcolsep}{6pt}
\begin{tabular}{lcc}
\toprule
Model & Acc. $\uparrow$ & Hallucination (\%)$\downarrow$ \\
\midrule
GPT-4o (zero-shot) & 61.22 & 7.00 \\
LLaVA-Med          & 87.75 & 2.00 \\
MedRAX             & 89.79 & 1.60 \\
\textbf{PASS}      & 93.50 & 1.60\\
\bottomrule
\end{tabular}
\caption{Performance on radiologist-verified safety-critical split from CAB-E. }
\label{tab:safety}
\end{table}

\begin{table}[t]
\centering\small
\begin{tabular}{lcc}
\toprule
\textbf{Configuration} & Acc. & $\Delta$Cost \\
\midrule
Full PASS                             & 91.22 & - \\
\quad– EarlyExit                      & 88.60 & 94.0 \\
\quad– Path-Rank Pretraining          & 87.86 & 8.9 \\
\quad– Expert-Guided Warm-up          & 88.89 & 9.5 \\
\bottomrule
\end{tabular}
\caption{Ablation study on CAB-E. $\Delta$Cost reports cost decrease relative to full PASS. }
\label{tab:ablation}
\end{table}

\noindent\textbf{Baselines.}
We compare PASS against four groups of methods:
\textbf{(1) general-purpose VLMs}, including GPT-4o~\cite{openai2024gpt4ocard}, the finetuned version of GPT-4o on the same training data of PASS, and its reasoning-augmented variants CoT~\cite{wei2022chain}, ComplexCoT~\cite{fu2023complexitybased}, and SC (CoT$\times$5)~\cite{zhang2023selfconsistency};
\textbf{(2) reasoning-centric VLMs}, o3-mini~\cite{o3-mini} paired with LLaVA-Med~\cite{li2023llava} as a visual captioning front-end due to its lack of image input;
\textbf{(3) medical/CXR-specialized VLMs}, LLaVA-Med and CheXagent~\cite{cxragent2024};
and \textbf{(4) agentic systems}, including the multimodal system MedRAX~\cite{fallahpour2025medrax} and originally single-modality methods (e.g., MaAS~\cite{maas2025}, AFlow~\cite{zhang2025aflow}), which we adapt to the multimodal setting by augmenting them with the same vision tools as PASS, with detailed results for these adapted agents reported in Appendix E.

\paragraph{Implementation details.}
We optimize the model using the AdamW algorithm, incorporating gradient clipping at 1.0 to ensure numerical stability, a weight decay of 0.01 to prevent overfitting, and a cosine learning rate schedule to facilitate smooth convergence. An entropy bonus of 0.01 is applied to encourage exploration and stabilize training. For RL updates, we employ forward-mode unrolling with a 5-step truncation to balance computational efficiency and gradient accuracy.

\begin{figure}[t]
 \centering
  \includegraphics[
    width=0.46\textwidth,
    height=0.23\textheight,
    keepaspectratio
  ]{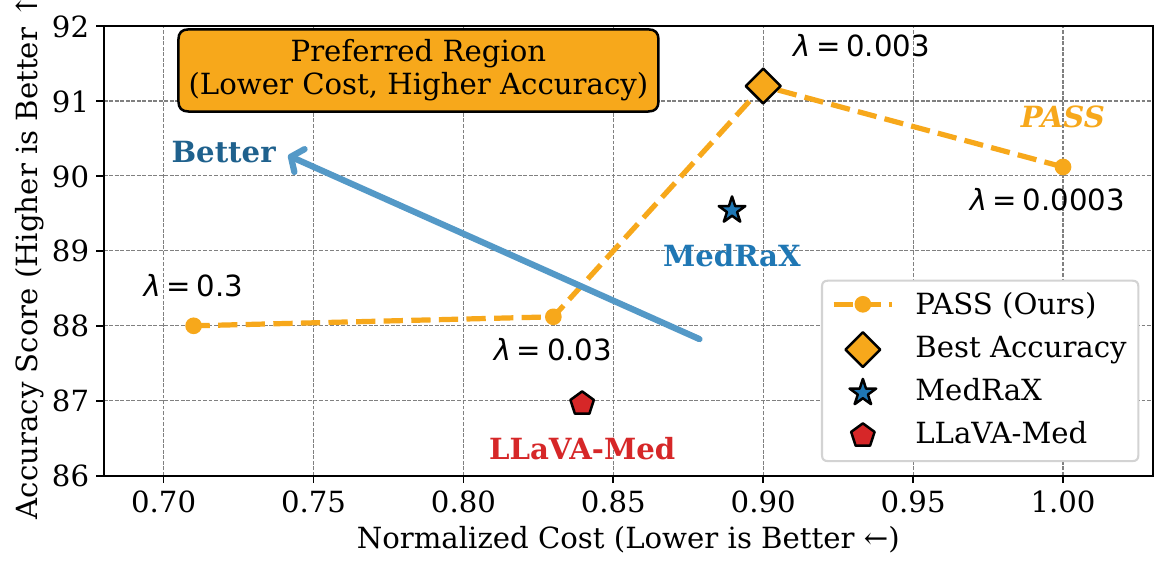}
 \caption{Cost-Accuracy Pareto Frontier analysis. Each orange point on the dashed frontier corresponds to a specific penalty weight ($\lambda$) configuration of PASS, enabling flexible cost–accuracy trade-offs at deployment. MedRAX and LLaVA-Med are plotted as additional points for comparison. Lower normalized inference cost and higher accuracy are preferred; the arrow indicates the desired direction toward the top-left preferred region.}
 \label{fig:cost_accuracy_frontier}
\end{figure}

%---------------------------------------------------------------
\subsection{Performance Analysis}
%---------------------------------------------------------------

Table~\ref{tab:main_results} presents the results on CAB-E, CAB-Standard, and SLAKE. PASS achieves an accuracy of 91.22, outperforming the strongest baseline MedRAX (89.54) by +1.68, surpassing CheXagent by +7.55 and LLaVA-Med by +4.26, demonstrating substantial improvement in diagnostic accuracy through probabilistic multi-tool reasoning. This suggests that adaptively sampled agentic trajectories, rather than single-pass VLMs or black-box agent planners, offer superior coverage and reliability on diverse CXR cases. 
We also observed that a specific version of GPT-4o that is finetuned on the same training dataset of PASS lags behind PASS, suggesting that probabilistic, query-dependent tool trajectories are the key factor, not merely domain-specific training.

PASS also achieves the highest LLM-J. score (84.28), METEOR (33.21), ROUGE-L score (31.49), and the second-best BLEU (8.51) among all strong baselines. This indicates that the answers provided by PASS align better with ground truth clinical solutions, validating the controller’s ability to coordinate image grounding, clinical reasoning, and textual fluency across multi-hop tool outputs.

\subsection{Latency and Cost Analysis.}

Table~\ref{tab:main_results} shows that while PASS exhibits higher latency than single-pass models like LLaVA-Med, this is a direct and strategic trade-off for its superior accuracy, driven by a more comprehensive reasoning process.
Figure~\ref{fig:cost_accuracy_frontier} illustrates the empirical cost–accuracy Pareto frontier of PASS by varying the penalty weight~$\lambda$, where the x-axis denotes normalized inference cost (relative to $\lambda = 0.0003$) and the y-axis reports accuracy. As $\lambda$ increases, PASS traverses a smooth frontier that substantially reduces cost with only modest accuracy degradation, exposing multiple deployment-ready operating points. The highest accuracy (91.2\%) is achieved at an intermediate setting $\lambda = 0.003$, where PASS outperforms MedRAX and LLaVA-Med by $1.66$ and $4.24$ absolute accuracy points, respectively, at comparable cost. For more aggressive cost-saving, larger $\lambda$ values (e.g., $\lambda = 0.03$) further reduce cost by roughly 20\% while still retaining around 88\% accuracy. Overall, PASS learns a well-structured frontier, enabling practitioners to tune $\lambda$ at deployment time to match latency and budget constraints without retraining.

%---------------------------------------------------------------
\subsection{Safety-Critical Subset Evaluation}
%---------------------------------------------------------------

On this safety-critical CAB-E subset, PASS achieves an accuracy of 93.50\%, surpassing MedRAX by 3.71 percentage points and LLaVA-Med by 5.75 percentage points. Notably, PASS and MedRAX share the lowest hallucination rate, representing a substantial improvement over the GPT-4o baseline and highlighting PASS's robustness in minimizing errors on safety-critical CXR cases. A blind human radiologist review further corroborates the superiority of PASS, with details provided in the Appendix. Taken together, these results underscore PASS's reliability in safety-critical clinical reasoning scenarios.

%---------------------------------------------------------------
\subsection{Ablation Study}
%---------------------------------------------------------------

Ablation results (Table \ref{tab:ablation}) confirm critical design choices: Removing early-exit causes a significant accuracy drop (from 91.22 to 88.60) and a 94\% relative cost decrease. Removing path-rank pretraining and warm-up also demonstrates their role in convergence acceleration and performance improvements.

\section{Related Work}

\paragraph{Tool-augmented LLMs.}
Tool use in LLMs has evolved from basic augmentation~\cite{schick2023toolformer,yao2023react,feng2025demograph} to modular agent frameworks~\cite{autogen2023,camel2023,chen2023agentverse,zhuge2024gptswarm} with specialized roles and communication. Yet, most rely on static or handcrafted workflows, limiting adaptability and efficiency in real-world deployment. Recent work begins to automate tool strategies and workflows via reinforcement learning or structured search~\cite{feng2025retool,zhang2025aflow}, but typically commits to a single, task-agnostic pipeline and offers little support for uncertainty-aware or dynamically adaptive inference.

\paragraph{Autonomous agent workflows.}
Recognizing the limitations of fixed pipelines, a new wave of research seeks to automate agentic system design. Prompt optimization~\cite{khattab2024,guo2024}, inter-agent communication tuning~\cite{gdesigner2024}, and modular profiling~\cite{chen2024autoagents} are key directions. Notably, MaAS~\cite{maas2025} introduces an agentic supernet that learns a distribution over multi-agent architectures and samples query-dependent workflows, improving accuracy–cost trade-offs and transferability beyond static designs. However, these approaches remain largely confined to text-only domains and offer limited interpretability and explicit uncertainty modeling, which is particularly problematic in high-stakes applications such as medicine.

\paragraph{Multimodal reasoning in medical AI.}
Multimodal foundation models (e.g., GPT-4V~\cite{liu2024systematic}, LLaVA-Med~\cite{li2023llava}, CheXagent~\cite{cxragent2024}) promise unified vision-language understanding and have shown zero-shot capabilities across radiological tasks. Still, they often hallucinate~\cite{eriksen2024use}, lack task specificity~\cite{cxragent2024}, and remain opaque. Domain-specific systems like MedRAX~\cite{fallahpour2025medrax} and MDAgents~\cite{kim2024mdagents} attempt to integrate medical tools with LLMs via ReAct-style~\cite{yao2023react} prompting, offering partial medical multimodal reasoning capabilities. Yet, their decision-making still largely relies on black-box LLMs, hindering real-world application due to critical concerns about trust and potential risks.

\paragraph{Safety and interpretability in clinical deployment.}
Clinical settings demand more than performance: they require transparency, controllability, and regulatory compliance~\cite{lundervold2019overview}. Beyond saliency-based explanations, methods like MedCoT~\cite{liu2024medcot} and BoxMed-RL~\cite{jing2025reason} leverage chain-of-thought or RL-enhanced generation to increase reliability. PASS extends these efforts with per-step, probability-annotated execution traces and interpretable early exits, allowing for post-hoc audits and fine-grained trust calibration, which are crucial features for safe medical AI deployment.

\section{Conclusion}

In this paper, we introduce PASS, the first multimodal framework to address the critical challenges of interpretability, adaptability, and efficiency in complex chest X-ray reasoning. Existing agentic systems are often limited by their black-box nature, poor integration of multimodal data, and rigid, inefficient workflows. PASS overcomes these limitations by leveraging a probabilistic controller to adaptively sample workflows from a multi-tool supernet, yielding decision paths annotated with transparent probabilities that are crucial for clinical trust and post-hoc audits. Our novel three-stage training strategy performs expert knowledge warm-up, contrastive path-ranking, and cost-aware reinforcement learning to optimize the performance-cost trade-off, balancing diagnostic accuracy with computational cost via a dynamic early-exit mechanism. Through extensive experiments on our newly curated \textsc{CAB-E} and other public benchmarks, we have demonstrated that PASS not only achieves superior accuracy over strong baselines but also provides interpretable and efficient reasoning. Ultimately, we believe that PASS represents a paradigm shift towards the next generation of multimodal, trustworthy, adaptive, and resource-aware agentic systems, grounded in medical reasoning yet potentially broadly applicable to other multimodal or high-stakes domains.

\paragraph{Limitations.}

PASS deliberately uses a fixed container set to ensure clinical safety and interpretability as a strategic trade-off between safety and flexibility. Future works will scale this robust foundation by expanding the supernet to new imaging types like MRI or CT, and enriching its agentic containers and tools, thereby further enhancing the diagnostic utility and adaptability of PASS.

\section*{Acknowledgements}

This work was supported in part by the Research Grants Council of Hong Kong (27206123, 17200125, C5055-24G, and T45-401/22-N), the Hong Kong Innovation and Technology Fund (GHP/318/22GD), the National Natural Science Foundation of China (No. 62201483), and Guangdong Natural Science Fund (No. 2024A1515011875).

\bibliography{aaai2026}

\onecolumn

\appendix
\section{Additional Details on Datasets}
\label{sec:appendixA}

\subsection{Additional Information on SLAKE and CAB-Standard Dataset}

We summarize the SLAKE and ChestAgentBench-standard datasets for multimodal CXR reasoning evaluation, detailed in Table~\ref{tab:dataset_summary}.

\subsubsection*{SLAKE Dataset}
SLAKE \cite{shao2021slake} is a public bilingual Med-VQA dataset with 642 images (CT: 282, MRI: 181, X-Ray: 179) from Medical Decathlon, NIH ChestXray-NIHCC, and CHAOS, covering head (140), neck (41), chest (219), abdomen (201), and pelvic cavity (41). Annotations include segmentation masks and bounding boxes for 12 diseases and 39 organs. It has 14,028 QA pairs (English/Chinese, vision-only/knowledge-based), a 5,232-triplet knowledge graph, and splits into 450/96/96 images (train/val/test). 

\subsubsection*{ChestAgentBench-standard Dataset}
The ChestAgentBench-Standard (CAB-Standard) dataset consists of 2,050 multiple-choice questions generated from 205 images selected from the Slake dataset, using the construction methodology of MedRaX \cite{fallahpour2025medrax}, which was originally based on Eurorad dataset. Slake was chosen over Eurorad due to its one-image-multiple-questions nature (instead of mutiple-image-multiple-questions style of Eurorad), and its superior compatibility with visual question answering tasks. This dataset provides a high-quality benchmark for evaluating AI systems in medical reasoning and decision-making.

CAB-Standard can be constructed by running the code scripts provided in the supplementary material.

\begin{table}[h]
\caption{Summary of SLAKE and ChestAgentBench-standard datasets.}
\label{tab:dataset_summary}
\centering
\small                       
\setlength{\tabcolsep}{4pt}  
\renewcommand{\arraystretch}{1.1}

\begin{tabularx}{\columnwidth}{>{\bfseries}l >{\centering\arraybackslash}X >{\centering\arraybackslash}X}
\toprule
\rowcolor[gray]{0.90}
Dataset & SLAKE & CAB-Standard \\
\midrule
\# Images          & 642   & 205 \\
\rowcolor[gray]{0.97}
\# QA Pairs        & 14,028 & 2,050 \\
Questions          & Vision-only, Knowledge-based & Detection, Classification, Localization, Comparison, Relationship, Diagnosis, Characterization \\
\rowcolor[gray]{0.97}
Language           & EN \& ZH & EN \\
Body Parts         & Head (140), Neck (41), Chest (219), Abdomen (201), Pelvic (41) & Chest \\
Modalities         & CT (282), MRI (181), X-Ray (179) & X-Ray \\
\rowcolor[gray]{0.97}
Annotations        & Masks, boxes (12 diseases, 39 organs) & Clinical details, findings \\
Source             & Medical Decathlon, NIH ChestXray, CHAOS & Slake \\
\rowcolor[gray]{0.97}
Split              & 450/96/96 (images) & 156/49 (images) \\
QA Type & Textual answer in around 1-3 words & Multiple-choice questions \\
\bottomrule
\end{tabularx}
\end{table}

\subsection{Additional Details on CAB-E Dataset}

To systematically evaluate agentic reasoning in safety-critical radiological settings, we introduce \textbf{ChesrAgentBench-Enhanced (CAB-E)}, a benchmark comprising 2,550 multi-step CXR cases designed to require complex, clinically grounded visual-language reasoning. Each case contains a CXR image, a free-form clinical query, and a detailed expert-written answer. Among these, 500 cases involve high-risk scenarios validated by a board-certified radiologist. Questions are generated using structured prompts that enforce 2–5 distinct analytical steps (e.g., localization, classification, cross-region comparison, and interpretation of clinical implications), while concealing tool names to prevent response leakage.

CAB-E is publicly available via the aforementioned URL for research purpose only.

To initialize our agent's controller, we conduct a supervised warm-up phase using ground-truth tool sequences retrieved by knowledge distillation from a domain knowledge-intensive model (i.e., GPT-4o). This phase is based on a fixed set of 500 warm-up samples, which are carefully curated with radiologist-in-the-loop feedback to ensure clinical relevance and accuracy. The radiologist feedback plays a crucial role in continuously guiding and refining the curation of these samples, enhancing their quality and alignment with domain-specific needs. During this warm-up, the controller is trained for 3 epochs to imitate the expert tool-use trajectories provided by GPT-4o, explicitly learning to predict the correct sequence of tools for a given query. 
This pre-training bootstraps the agent with foundational knowledge, which significantly stabilizes and accelerates the subsequent, more complex reinforcement learning phase.

CAB-E emphasizes interpretability and modular decision-making, with each case accompanied by a list of required tools (e.g., segmentation, classification, VQA, report generation) used to guide model reasoning. Evaluation is conducted using a rubric-based scoring protocol across four criteria: \textit{correctness, completeness, relevance, and coherence}.

A concrete example is shown below:

\begin{promptbox}
Image: `images/source0001.jpg'  \\

Question: Based on the chest X-ray provided, identify and locate any abnormalities within the pleural spaces, particularly focusing on the right lung and the lower left chest region. Determine if there is evidence of pleural effusion and evaluate its extent in these areas. Establish the relationship between the effusion in the right lung and the lower left chest region, explaining any potential underlying causes and clinical significance of these findings.  \\

Answer: The chest X-ray reveals a pleural effusion affecting the right lung and extending into the lower left chest region. The effusion is characterized by fluid accumulation in the pleural space, particularly notable on the right side. The relationship between the findings suggests that the effusion is more pronounced on the right, potentially indicating a unilateral process or a difference in fluid dynamics between the lungs. The clinical significance of these findings could be related to conditions such as infection, heart failure, or malignancy, which need to be further investigated through clinical correlation and possibly additional diagnostic tests.  \\

Required Tools: `ChestXRaySegment → ChestXRayClassify → VQAnalyze → ChestXRayReport → EarlyStop'
\end{promptbox}

\subsection{Safety-critical Split in the CAB-E Datasets}
\paragraph{Selection Criteria.} The 500 safety-critical instances are constructed alongside the rest samples in the CAB-E dataset, following the same pipeline, only with additional prompt guidance on safety-critical aspects. 

We attach the prompt in Appendix D for your reference.

These cases were sampled from a broad pool of candidate CXRs in the SLAKE dataset to ensure coverage of common failure modes, focusing on high-risk diagnostic errors.

\paragraph{Inter-Rater Agreement.} The current annotations were provided by a single board-certified radiologist. While inter-rater agreement is not applicable in this setting, we plan to involve multiple radiologists in future work to enable agreement analysis and strengthen evaluation reliability.

\paragraph{Failure Mode Analysis.}
To quantitatively understand the model's failure modes, we performed a granular error analysis on all predictions that achieved an accuracy below 0.9 when compared against the ground truth. We developed a four-category classification scheme to move beyond simple error detection and pinpoint the specific nature of each failure. The categories were: (1) \textit{Factual Fabrication / Hallucination}, where the model generated information unsupported by the ground truth; (2) \textit{Omission of Primary Findings}, referring to the failure to report core diagnostic conclusions; (3) \textit{Omission of Finding Attributes}, where key descriptors like location or severity were missing; and (4) \textit{Omission of Clinical Implications}, where the model failed to address patient management aspects of the query. The distribution of these failure modes is summarized in Table~\ref{tab:failure_mode_distribution}.
\begin{table}[h!]
\centering\footnotesize
\caption{Failure mode distribution on the CAB-E safety subset (accuracy lower than 0.9).}
\label{tab:failure_mode_distribution}
\begin{tabular}{lr}
\toprule
\textbf{Failure Mode}                     & \textbf{Proportion} \\
\midrule
Omission of Clinical Implications         & 68.8\% \\
Omission of Primary Findings              & 16.8\% \\
Omission of Finding Attributes            & 12.8\% \\
Factual Fabrication / Hallucination       & 1.6\%  \\
\midrule
\textbf{Total}                            & \textbf{100.0\%} \\
\bottomrule
\end{tabular}
\end{table}

\paragraph{Safety Aspect Categorization.} We categorized safety-critical findings into four aspects: (1) life-threatening anatomical abnormalities, (2) critical positioning of medical devices, (3) urgent systemic conditions, and (4) acute cardiovascular or respiratory emergencies. Table~\ref{tab:safety_aspect_distribution} summarizes the distribution of these aspects within the subset.

\begin{table}[h]
\centering\footnotesize
\caption{Distribution of safety-critical aspects within the safety-critical subset of CAB-E.}
\label{tab:safety_aspect_distribution}
\begin{tabular}{lcc}
\toprule
\textbf{Safety Aspect} & \textbf{Count} & \textbf{Proportion} \\
\midrule
Life-threatening anatomical abnormalities & 296 & 59.2\% \\
Urgent systemic conditions (e.g., ARDS, sepsis) & 149 & 29.8\% \\
Acute cardiovascular or respiratory emergencies & 51 & 10.2\% \\
Critical positioning of medical devices (e.g., misplaced tubes or lines) & 4 & 0.8\% \\
\midrule
\textbf{Total} & 500 & 100\% \\
\bottomrule
\end{tabular}
\end{table}

\section{Additional Details on Evaluation Metrics}
\label{sec:appendixB}

\subsection{Quantitative Evaluation Metrics}

We define the evaluation metrics for assessing model performance.

\begin{itemize}[itemsep=2pt, parsep=2pt]
    \item \textbf{Accuracy}: The fraction of correctly predicted labels over all ground-truth instances. For open-ended QA, we use an LLM-as-judge score (LLM.J) $s \in [0,1]$ between the model prediction and the reference answer, and then binarize it: cases with $s > 0.5$ are counted as correct and those with $s \le 0.5$ as incorrect.
    \item \textbf{Semantic/Embedding Similarity}: We quantify the semantic alignment between the generated response and the reference using the \texttt{all-MiniLM-L6-v2} model from Sentence-Transformers. We compute the cosine similarity between the prediction embedding $\mathbf{e}_p$ and reference embedding $\mathbf{e}_r$, and rescale the result to the range $[0,1]$:
\[
\text{Sim} = \frac{1}{2} \left( \frac{\mathbf{e}_p \cdot \mathbf{e}_r}{\|\mathbf{e}_p\| \|\mathbf{e}_r\|} + 1 \right).
\]
This rescaling ensures that the metric is strictly non-negative, where 1 indicates identical semantic meaning and 0 indicates opposite meaning.
    \item \textbf{BLEU}: Bilingual Evaluation Understudy score, measuring the similarity between predicted and reference text using n-gram precision, typically weighted as
    \[
    \text{BLEU} = \text{BP} \cdot \exp\left(\sum_{n=1}^{N} w_n \log p_n\right),
    \]
    where BP is a brevity penalty, $p_n$ is the n-gram precision, and $w_n$ are weights (often $w_n = 1/N$).
    \item \textbf{Token}: The average number of tokens (words or subwords) in model outputs, reflecting computational efficiency. The latency comparison is fair since the OpenAI API-based baselines are served by highly optimized, large-scale data center clusters that can offer comparable latency with our single research GPU, meaning our setup does not give PASS an unfair advantage.
    \item \textbf{Latency}: The average time (in seconds) for the model to generate a response, measuring inference speed.

    \item \textbf{Hallucination}: The fraction of generated outputs containing factually incorrect or unsupported information, assessed via manual annotation, supplemented by automated fact-checking via strong language models. The automated fact-checking employs the following prompt for validation:
    \begin{promptbox}
        You are an expert in medical AI diagnostics. Your task is to evaluate a diagnostic model's prediction and determine whether it exhibits hallucination.\\
Here is the definition of hallucination:\\
Factual Fabrication / Hallucination: The model's prediction contains medical findings that are completely made up and are NOT supported by the ground truth.\\
Example: Ground Truth mentions only atelectasis, but the Prediction claims there is a ``large pulmonary nodule".\\

Analyze the following case and determine whether the model's response is a hallucination or not.\\

Context for Analysis:\\

Query: ``{entry[`query']}"\\
Ground Truth: ``{entry[`ground\_truth']}"\\
Model's Prediction: ``{entry[`prediction']}"\\
Your response MUST be either:\\

Yes (if hallucination is present)\\
No (if hallucination is NOT present)\\
    \end{promptbox}
    \item \textbf{LLM-as-a-Judge Score (LLM-J.)}
    \paragraph{Prompt Template for Assessing LLM-as-a-Judge Score}

\begin{promptbox}
    
You are an expert evaluator assessing the quality of generated answers against the ground truth. 
Evaluate text2 (Model Generated Answer) against text1 (Ground Truth Answer) using these criteria:\\

1. Correctness: Accuracy of the information.
2. Completeness: Coverage of key points from text1.
3. Relevance: Pertinence to the question/context.
4. Coherence: Clarity and logical flow.\\

SCORING RUBRIC:
- 1.00 (Clinically Correct): The answer is clinically correct and captures all major key points from the ground truth. Minor differences in phrasing or omission of trivial details are acceptable.
- 0.90 (Mostly Correct): Generally correct, covers most key points, with minor omissions or inaccuracies.
- 0.70 (Partially Correct): Captures some key points, generally relevant, but has notable omissions or inaccuracies.
- 0.50 (Partially Incorrect): Limited understanding with major errors or irrelevant parts.
- 0.00 (Incorrect): Completely incorrect, irrelevant, or refuses to answer.\\

Evaluate based on the above criteria and provide a single numeric score: 0.0, 0.5, 0.7, 0.9, or 1.0. Respond with ONLY the numeric score.\\

Ground Truth Answer: \{text1\}\\

Model Generated Answer: \{text2\}
\end{promptbox}

    \item \textbf{Normalized Inference Cost}: We quantify inference cost as the weighted sum of LLM inference tokens (via APIs) and FLOPs consumed by locally executed tools over the agentic workflow. The conversion between token usage and FLOPs follows \citet{kaplan2020scaling}. For ease of comparison, all reported costs are normalized by the cost of PASS at $\lambda = 0.0003$.
\end{itemize}

\subsection{Qualitative Case Study: Blind Human Radiologist Expert Evaluation}

To qualitatively assess the clinical behavior of PASS beyond automatic metrics, we conducted a blinded reader study with a board-certified radiologist on a subset of \textsc{CAB-E}. We randomly sampled 100 de-identified image–question–answer cases spanning a range of findings, difficulty levels, and safety-critical scenarios. 

A radiologist with 7 years of experience in chest and abdominal X-ray interpretation independently reviewed each CXR, the corresponding clinical question, and the model-generated answer from seven systems (GPT-4o, GPT-4o finetuned on PASS's domain-specific data, LLaVA-Med, o3-mini+LLaVA-Med, MedRAX, CheXagent, and PASS). For every response, the radiologist: (i) assigned an overall rating ranging from 1 to 5 relative to the reference report and standard clinical practice; (ii) flagged whether the answer was \emph{clinically unsafe or misleading} (e.g., an incorrect high-risk diagnosis, inappropriate management, or hallucinated critical findings); and (iii) provided brief free-text expert commentary on notable strengths and failure modes. All cases were fully de-identified in accordance with institutional policies and applicable regulations. The radiologist was blinded to model identity and only provided with anonymized labels (“Model A,” “Model B,” “Model C,” etc.).

Across the 100-case subset, PASS achieved the highest proportion of fully correct and clinically safe answers and the lowest rate of responses flagged as unsafe or misleading among all compared systems. The expert highlighted PASS’s tendency to remain strictly grounded in the information provided by upstream tools and the case description, avoiding speculative lesion measurements or unsupported alternative diagnoses. In contrast, several baselines occasionally produced overconfident but incorrect or hallucinatory content (e.g., invented lesion sizes or specific fungal etiologies). This pattern is consistent with PASS’s design as a probabilistic, tool-aware controller that favors calibrated, evidence-aligned reasoning over aggressive extrapolation.

Below, we present a representative qualitative case to illustrate these differences.

\vspace{1em}

\noindent
\begin{promptbox}

\textbf{Original Question:} Evaluate the chest X-ray to identify and count the areas of consolidation due to pneumonia in both lungs. Compare the extent of these consolidations between the left and right lungs. Based on the comparison, diagnose the condition, and suggest a rationale for any recommended treatments or interventions.

\begin{center}
    \includegraphics[width=0.3\linewidth]{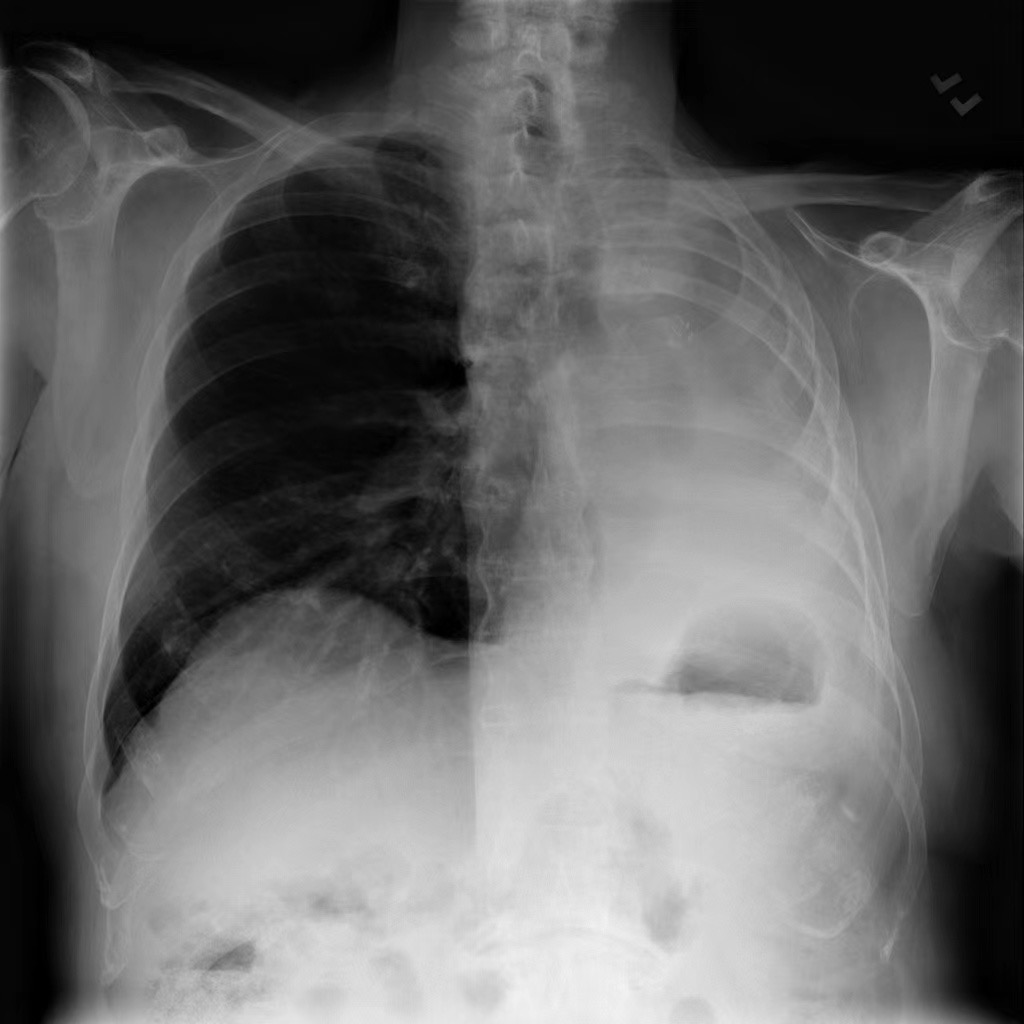}
    
    \vspace{8pt}
    {\small\textit{Chest X-ray image used in the medical reasoning case study.}}
\end{center}

\vspace{10pt}
\end{promptbox}
\vspace{10pt}

\noindent \textbf{Model Responses and Expert Evaluations:}

\begin{promptbox}

\textbf{GPT-4o (Finetuned on domain-specific dataset used by PASS)} \\
          \textbf{Response:} The chest X-ray reveals multiple opacities in the left lung's right side. Among these, one opacity is notably larger than the others, indicating a dominant lesion. The presence of these opacities suggests a condition such as pneumonia or a lung mass. Given the characteristics and distribution, pneumonia is a likely diagnosis. The recommended treatment plan would involve antibiotics to address the infection, along with supportive care to manage symptoms and improve lung function. \\
          \textbf{Score:} 3 \\
          \textbf{Expert Commentary:} This response is mostly accurate.It covers key findings, reasonable diagnoses, and treatment suggestions. However, it does not clearly mention important imaging features supporting a pneumonia diagnosis, such as ``fluid accumulation" and "consolidation."
This weakens the rigor of its diagnostic reasoning.
 \end{promptbox}

\begin{promptbox}
\textbf{LLaVa-Med} \\
    \textbf{Response:} The chest X-ray shows multiple opacities in the left lung's right side. The largest one measures approximately 2.5 cm. Based on these findings, the condition affecting the lung is likely a fungal infection, specifically aspergillosis. Aspergillosis is a fungal infection caused by the Aspergillus species, which can affect various parts of the body, including the lungs. The presence of multiple opacities and the largest one measuring 2.5 cm are indicative of this condition. \\
    \textbf{Score:} 1 \\
    \textbf{Expert Commentary:} This response deviates far from the standard answer. It misdiagnoses aspergillosis. It invents precise details (``2.5 cm”) but omits key signs like fluid accumulation or consolidation. It also lacks any treatment suggestions. Overall, its credibility and clinical usefulness are very low.
\end{promptbox}

\begin{promptbox}

\textbf{o3-mini + LLaVA-Med as visual tool} \\
          \textbf{Response:}  

Number of Opacities and Dominant Lesion:  
The chest X-ray description indicates the presence of multiple opacities in the left lung's right side. While the exact count of these opacities is not specified, the reference to ``multiple'' suggests more than two distinct areas. Among these, the largest opacity measures approximately 2.5 cm, making it the dominant lesion compared to the smaller opacities.

Diagnosis:  
Based on the described findings, the most likely diagnosis is aspergillosis, a fungal infection caused by Aspergillus species. This condition is characterized by the formation of lung abscesses, which appear as opacities on imaging studies. The presence of multiple opacities, particularly with a sizable dominant lesion of 2.5 cm, is indicative of this invasive fungal infection.

Rationale:  
\begin{itemize}[leftmargin=1.5em]
    \item Multiple Opacities: Suggestive of widespread infection or abscess formation within the lung tissue.
    \item Dominant Lesion (2.5 cm): A larger opacity may represent a significant abscess or localized area of intense infection, consistent with aspergillosis.
    \item Fungal Etiology: Aspergillosis often occurs in individuals with compromised immune systems or underlying lung conditions, leading to the formation of abscesses observable on X-rays.
\end{itemize}

Recommended Treatment Plan:  
\begin{enumerate}[leftmargin=1.8em]
    \item Antifungal Therapy:
          \begin{itemize}[leftmargin=1.2em]
              \item First-Line Treatment: Initiate voriconazole, which is the preferred antifungal agent for invasive aspergillosis.
              \item Alternative Options: If voriconazole is contraindicated or not tolerated, alternatives include isavuconazole or liposomal amphotericin B.
          \end{itemize}
    \item Supportive Care:
          \begin{itemize}[leftmargin=1.2em]
              \item Monitor Respiratory Status: Regular assessment of respiratory function and oxygenation.
              \item Manage Underlying Conditions: Address any predisposing factors, such as immunosuppression or chronic lung diseases.
          \end{itemize}
    \item Follow-Up Imaging:
          \begin{itemize}[leftmargin=1.2em]
              \item Assess Treatment Response: Repeat chest imaging (e.g., X-ray or CT scan) to evaluate the reduction in size and number of opacities, indicating improvement.
          \end{itemize}
    \item Consultation with Specialists:
          \begin{itemize}[leftmargin=1.2em]
              \item Infectious Disease: For guidance on antifungal therapy and management of complications.
              \item Pulmonology or Thoracic Surgery: If there is no response to medical therapy, surgical intervention may be considered to drain abscesses.
          \end{itemize}
    \item Monitoring for Side Effects:
          \begin{itemize}[leftmargin=1.2em]
              \item Medication Tolerance: Regular monitoring for adverse effects of antifungal agents, such as liver function tests for voriconazole.
          \end{itemize}
\end{enumerate}

Note: It is essential to consult a healthcare professional for an accurate diagnosis and tailored treatment plan.\\
          \textbf{Score:} 2 \\
          \textbf{Expert Commentary:} This response has a rigorous structure and rich clinical details. However, its diagnostic direction deviates significantly from the correct answer. It misdiagnosed aspergillosis instead of pneumonia. It also invented imaging details not provided (e.g., ``2.5 cm"). This significantly lowers its overall credibility and clinical plausibility.

\end{promptbox}

\begin{promptbox}
    \textbf{MedRAX} \\
          \textbf{Response:} Upon examination of the chest X-ray, multiple opacities are present in the left lung's right side, consistent with pneumonia. These opacities suggest fluid accumulation, typical in infectious lung conditions. The comparison reveals that one opacity is significantly larger, indicating a potential area of greater consolidation. The diagnosis is pneumonia, specifically affecting the left lung, which aligns with the increased opacities and fluid signs. The recommended treatment plan includes antibiotics to address the bacterial infection and possibly supportive care, such as oxygen therapy, to manage respiratory distress. \\
          \textbf{Score:} 4 \\
          \textbf{Expert Commentary:} This is a strong, clinically coherent response. It correctly recognizes multiple opacities in the left lung, links the presence of a larger opacity to greater consolidation, and reasonably concludes pneumonia as the working diagnosis. The explanation appropriately mentions fluid accumulation and consolidation and proposes a suitable management plan with antibiotics and supportive care such as oxygen therapy for respiratory distress. However, the answer remains somewhat general in two aspects. First, it does not truly enumerate the opacities as requested (e.g., approximate number or distribution) and instead only describes them as `multiple.’ Second, it could briefly acknowledge the need for clinical correlation or consider close differentials (such as other causes of localized consolidation) before committing to pneumonia. With these additions, it would reach an excellent, more comprehensive level. Overall, it is a reliable and largely well-reasoned interpretation.
\end{promptbox}

\begin{promptbox}
         \textbf{PASS (ours)} \\
          \textbf{Response:} Based on the provided information, the chest X-ray indicates the presence of opacities in the left lung's right side, which are characteristic of pneumonia. However, the specific number of opacities and their sizes are not detailed in the information given. Therefore, I cannot enumerate the opacities or determine if there is a dominant one.
          
          Given the conclusion of pneumonia, the diagnosis is supported by the typical appearance of opacities on the X-ray, which suggest an infection in the lung tissue. The treatment plan for pneumonia generally includes antibiotics to target the causative bacteria, along with supportive care such as hydration, rest, and possibly bronchodilators if wheezing is present. If the pneumonia is viral, antiviral medications may be considered, but supportive care remains the primary focus. Further evaluation may be necessary to tailor the treatment based on the patient's specific condition and response to initial therapy. \\
          \textbf{Score:} 4 \\
          \textbf{Expert Commentary:} 
          This response provides a correct diagnosis of pneumonia and an appropriate, clinically sound treatment plan. Importantly, it remains strictly grounded in the information actually available from the case and the underlying investigation results, avoiding fabrication of unprovided details such as exact lesion size or precise opacity counts. This reflects good investigation and a well-calibrated reasoning procedure that prioritizes fidelity to the data over speculative imaging descriptions, in contrast to other answers that introduce spurious measurements or alternative pathologies. Although it is conservative in not enumerating the opacities, it still correctly links the described opacities to an infectious process and outlines a clear management strategy. Overall, it demonstrates a reliable and fact-grounded reasoning procedure, low hallucination risk, and strong alignment with the true clinical context.

\end{promptbox}

\section{Implementation Details}
\label{sec:appendixC}

\paragraph{Temperature Annealing.}
To balance exploration and stability during multi-agent workflow sampling, we adopt a linear temperature annealing schedule. Specifically, the controller's sampling temperature is initialized at 1.0 and linearly decays to 0.2 over the course of training. This encourages diverse trajectory exploration in early stages while gradually shifting toward deterministic and high-confidence tool selections as training progresses.

\paragraph{Configurations.}
We train the controller using the AdamW optimizer with a learning rate of 5e\textminus5, weight decay of 0.01, and gradient clipping at 1.0 to ensure training stability. A cosine learning rate decay schedule is applied to facilitate smooth convergence. To encourage action diversity and prevent premature convergence, an entropy regularization term with weight 0.01 is added to the policy loss. Reinforcement learning is performed via forward-mode unrolling with a truncation horizon of 5 steps, providing a trade-off between computational cost and gradient quality. All experiments are conducted on a single NVIDIA H800 GPU with 80GB of memory. The batch size is set to 32, and model checkpoints are saved every epoch for evaluation and ablation analysis.

\section{Additional Details on Specialized Tools}
\label{sec:appendixD}

We provide additional details on the key tool models encapsulated within the containers of the agentic supernet. The tool set can be flexibly customized by end users or deploying institutions to accommodate specific application requirements, evolving model versions, or privacy considerations. As the tools are confined within a pre-defined container interface, such customization and updates can be performed in a plug-and-play manner, without requiring any retraining of the PASS controller model.

\begin{itemize}[itemsep=2pt, parsep=2pt]
    \item \textbf{Classification}: This tool classifies chest X-ray images for 18 pathologies using a pre-trained DenseNet model. It predicts probabilities (0 to 1) for conditions like Atelectasis, Cardiomegaly, Consolidation, Edema, Effusion, Emphysema, Enlarged Cardiomediastinum, Fibrosis, Fracture, Hernia, Infiltration, Lung Lesion, Lung Opacity, Mass, Nodule, Pleural Thickening, Pneumonia, and Pneumothorax. Higher values indicate greater likelihood of the condition's presence, with output as a dictionary of pathologies and probabilities. The URL of the tool is: \url{https://huggingface.co/torchxrayvision/densenet121-res224-all}. 
    \item \textbf{Segmentation}: This tool segments chest X-ray images to identify and outline specific anatomical structures. It supports segmentation of organs such as Left/Right Clavicle, Left/Right Scapula, Left/Right Lung, Left/Right Hilus Pulmonis, Heart, Aorta, Facies Diaphragmatica, Mediastinum, Weasand, and Spine. Users can specify a list of organs or segment all available ones by default, with input being the image path and optional organ list. The URL of the tool is: \url{https://github.com/mlmed/torchxrayvision}.
    \item \textbf{Grounding}: This tool grounds medical findings in chest X-ray images using the MAIRA-2 model. It locates specific phrases like 'Pleural effusion' or 'Cardiomegaly' and returns bounding box coordinates (normalized 0-1), a visualization, and confidence scores. Input includes the image path, phrase, and optional parameters like max-new-tokens for processing. The URL of the tool is: \url{https://huggingface.co/microsoft/maira-2}.
    \item \textbf{VisualQA}: This tool leverages CheXagent for comprehensive chest X-ray analysis, supporting tasks like visual question answering, report generation, and abnormality detection. It accepts image paths and natural language prompts to provide detailed clinical interpretations or anatomical descriptions. The output varies based on the requested analysis, offering versatile diagnostic support. The URL for the tool is: \url{https://github.com/Stanford-AIMI/CheXagent}.
    \item \textbf{LLaVA-Med}: This tool uses a large language model fine-tuned on medical images to answer medical visual questions. It processes both image-based queries and general medical questions, providing accurate responses tailored to the input. The tool is ideal for detailed medical inquiries requiring image context or standalone medical knowledge. The URL of the tool is: \url{https://github.com/microsoft/LLaVA-Med}.
    \item \textbf{Report Generation}: This tool generates structured chest X-ray reports with detailed findings and impression summaries using Vision-Encoder-Decoder models trained on CheXpert and MIMIC-CXR datasets. It takes a chest X-ray image path as input and outputs a radiology report in standard format. The report includes comprehensive observations and concise clinical conclusions. The URL of the tool is: \url{https://huggingface.co/IAMJB/chexpert-mimic-cxr-findings-baseline}.
    \item \textbf{Guideline Lookup} This agentic search tool looks for accurate, updated, professional and domain-specific information to empower and verify the CXR reasoning. The URL of the tool is \url{https://github.com/tavily-ai}.

\end{itemize}

\section{Additional Details on Adapted Textual Agentic Planners}
\label{sec:AppendixE}

\begin{table}[h]
\centering\footnotesize
\caption{Performance of originally text-only agentic planners augmented with the same visual tools as \textbf{PASS} on the \textsc{CAB-E} benchmark. Higher is better (↑) except latency (↓, seconds). }
\label{tab:text-planners}
\begin{tabular}{lccccccc}
\toprule
\textbf{Method} & \textbf{Acc.\,$\uparrow$} & \textbf{LLM-J.\,$\uparrow$} & \textbf{BLEU\,$\uparrow$} & \textbf{METEOR\,$\uparrow$} & \textbf{ROUGE-L\,$\uparrow$} & \textbf{Sim.\,$\uparrow$} & \textbf{Lat.\,$\downarrow$} \\
\midrule
AFlow + visual tools          & 64.29   & 58.10   & 3.29   & \textbf{33.50}   & 22.63   & 77.51   & \textbf{12.32}   \\
MaAS + visual tools           & 72.75   & 67.35   & 5.92   & 24.95   & 28.68   & 83.67   & 28.01   \\
MedRAX          & 89.54 & 76.94 & 5.56 & 32.84 & 27.11 & 88.69 & 17.44 \\
\midrule
\textbf{PASS (ours)}          & \textbf{91.22} & \textbf{84.28} & \textbf{8.51} & 33.21 & \textbf{31.49} & \textbf{90.16} & 22.06 \\
\bottomrule
\end{tabular}
\end{table}

To ensure that PASS's gains are architectural rather than data- or model-specific, we also consider SOTA originally text-first agentic planners adapted to our setting.

Recent planner-search methods such as MaAS~\cite{maas2025} and AFlow~\cite{zhang2025aflow} optimize over text-domain agentic architectures for benchmarks in math, coding, web tools, and games. For a fair comparison on CXR reasoning, we keep each planner’s original search objectives, modules, and hyperparameters intact, and only expose a unified tool set implementing PASS’s visual functions. In other words, these planners still perform planning and routing purely in text, but can utilize our vision tools as callable modules. This design isolates the effect of the planner’s search strategy from differences in visual tooling, allowing us to specifically probe the benefit of PASS’s probabilistic controller.

Concretely, MaAS optimizes a distribution over agentic architectures for math and coding problems; and AFlow searches over code-represented workflows. Both of them are originally designed and evaluated for single-modality, non-medical tasks.

Under this adaptation, the agentic planners remain competitive but consistently underperform PASS on \textsc{CAB-E} across accuracy and safety-aligned metrics, while incurring comparable or higher tool costs. We attribute this gap to (i) their text-centric design, which lacks mechanisms for joint reasoning over visual and clinical signals, and (ii) the absence of native uncertainty-aware routing over visual evidence, which is explicitly modeled in PASS. Detailed results and ablations for these adapted planners are presented in Table~\ref{tab:text-planners}.

\section{Additional Details on Prompt Templates}
\label{sec:appendixF}

\paragraph{Prompt Template for Constructing Question-Answer Pairs in CAB-E Dataset}
For the medical X-ray question generation benchmark, we provide concrete prompt templates following our prompting strategy for both free-form and multiple-choice questions.

The following prompt is for generating free-form clinical questions that require detailed analysis and multi-step reasoning about chest X-ray images:

\begin{promptbox}
You must follow these guidelines: \\

        1. The question must focus on **one specific medical inquiry** related to chest X-rays. Avoid multiple sub-questions in the question itself.\\
        - It should require analyzing a chest X-ray to derive clinical insights.
        - The question must have a clear, verifiable answer.\\

        2. The question must be **clinically relevant** and require multi-step reasoning, such as:\\
        - Identifying specific structures or abnormalities.\\
        - Classifying findings and interpreting clinical significance.\\
        - Highlighting or comparing regions of interest.\\

        3. Ensure the analysis follows a **logical progression**. For example:\\
        - Step 1: Identify and segment relevant structures.\\
        - Step 2: Classify and describe abnormalities.\\
        - Step 3: Derive clinical interpretations related to the findings.\\

        4. When choosing tools:\\
        - Select only the tools necessary to answer the question logically and completely.\\

        **IMPORTANT REQUIREMENTS**:\\
        - The question must NOT include references to case IDs, file names, or software tools.\\
        - Do NOT mention specific tools in the question or answer.\\
        - The focus must be on medical reasoning, NOT technical instructions.\\
        - The answer must be based entirely and strictly on the provided clinical case (\{self.case\_content\}). Do NOT include any assumptions or content beyond the given case details.\\

        Format your response as follows:\\

        THOUGHTS: [Break down the reasoning process into clear steps and specify which tools are needed for each step, with justification.]\\

        QUESTION: [Write a single, focused clinical open-ended question. Avoid including any tool references.]\\

        REQUIRED\_TOOLS: [List 2-5 tools from the available tools in SEQUENCE to answer the question.]\\

        EXPLANATION: [Briefly explain why the tools are needed and how they work together to solve the question.]\\

        ANSWER: [Provide a detailed medical answer with findings and interpretation. The answer must be strictly based on \{self.case\_content\} without any additional assumptions. Do NOT mention tools in the answer.]\\

        **Available tools** (to be listed in REQUIRED\_TOOLS only):\\
        - ChestXRayClassify: Classifies X-rays for various conditions.\\
        - ChestXRaySegment: Segments different parts of chest X-rays.\\
        - ChestXRayReport: Generates medical reports from X-rays.\\
        - VQAnalyze: Answers visual questions about X-rays.\\
        - LlaVAMed: Understands and answers general medical image visual questions.\\
        - GuidelineLookUp: Searches for medical guidelines or other reliable knowledge sources.\\

        Your task: Create a \{self.difficulty\} \{self.type\} clinical question that integrates the following:\\

        \{category\_descriptions\}\\

        based on the following clinical case:\\

        \{self.case\_content\}\\

        Focus on requiring the agent to derive findings from the chest X-ray itself. Avoid providing explicit information about the case in the question.\\
\end{promptbox}

\paragraph{System Prompt for the Safety-Critical CAB-E Subset}
For the safety-critical subset of CAB-E, we use a dedicated system prompt to generate high-risk clinical scenarios. The prompt instructs the model to focus on specific life-threatening pathologies and to frame questions within a context of clinical urgency, ensuring the resulting cases test reasoning in high-stakes situations.
\begin{promptbox}
You are embodying a senior radiologist specializing in emergency and critical care. Your task is to create a benchmark question that tests an AI's ability to identify life-threatening or high-risk findings on a chest X-ray.\\

Your primary directive is to focus exclusively on scenarios where a missed or incorrect finding could lead to severe patient harm or immediate clinical crisis.\\

Examples of such critical findings include:\\
- Pneumothorax (especially tension pneumothorax)\\
- Misplaced lines or tubes (e.g., endotracheal tube in the esophagus, central line in an artery)\\
- Acute aortic dissection or aneurysm\\
- Pulmonary embolism\\
- Severe pneumonia or ARDS\\
- Pericardial effusion leading to tamponade\\

CRITICAL INSTRUCTION: In your `THOUGHTS' section, you must begin by explicitly stating *why* the chosen case is safety-critical. If the provided case does not contain a clear, high-risk finding, you should state that and still attempt to frame the most critical question possible.\\

AVOID generating questions about routine, non-urgent findings like stable nodules, old fractures, or mild degenerative changes. The question's focus must be on urgency and potential for immediate intervention.\\
\end{promptbox}

\paragraph{Prompt Template for Container Descriptions}
The following prompt is for embedding generation of agentic containers, providing detailed descriptions of each tool's capabilities and functions in chest X-ray analysis. These descriptions serve as semantic anchors for the controller to understand the purpose and functionality of each available medical specialized tool sets:
\begin{promptbox}
`ChestXRayClassify': ``Analyze chest X-rays and identify up to 18 possible pathologies, such as atelectasis, cardiomegaly, etc. Returns the probability of each disease.",\\

`ChestXRaySegment': ``Segment chest X-rays into different anatomical structures, such as left and right lungs, heart, aorta, etc., and provide the area and location information of each organ.",\\

`ChestXRayReport': ``Generate a detailed medical report based on the chest X-ray, including findings and impression sections, similar to a radiologist's report.",\\

`VQAnalyze': ``Answer specific medical questions about chest X-rays, combining visual understanding and medical knowledge. Can answer questions about lesions, structures, or diagnoses.",\\

`GroundFindings': ``Locate specific medical findings or abnormal regions on the chest X-ray, mapping text-described lesions to specific locations in the image.",\\

`EarlyStop': ``Complete the current analysis process and do not execute subsequent operators; used when there is already enough information to answer the query.",\\

`LlaVAMed': ``Use the LLaVA-Med model for general medical image visual question answering, capable of understanding and answering a wide range of questions about the input image."\\

`GuidelineLookup': ``Use the agentic search tool to look for accurate, updated, professional, and domain-specific information to empower and verify the CXR reasoning.''
\end{promptbox}

\paragraph{Prompt Template for GPT Response Aggregation}

The following prompt is for generating concise medical responses based on information aggregated from multiple tool outputs. It structures the collected data from X-ray image analysis tools into a coherent format that guides the language model to synthesize findings and produce answers to medical queries:
\begin{promptbox}
Context: The following information was gathered by an AI agent trying to answer a medical query about an image.\\

Original Query: \{query\}\\

Gathered Information:
\{gathered\_info\}\\

Based ONLY on the Original Query and the Gathered Information provided above, please provide a concise, factual answer to the Original Query. If the information is insufficient, state that.\\
Answer:\{\}
\end{promptbox}

\end{document}